# GAUSS: A Workload-Aware Analytical and Monte-Carlo Simulator for LLM Inference Serving

Srikant Bharadwaj, Ankur Mallick, Spyridon Mastorakis, Anjaly Parayil, Dhruv Deshmukh, Victor Ruehle
Microsoft

*Technical Report*

**Abstract**

*Large language model (LLM) inference has become an operational workload that fleets must provision, monitor, and bill against latency service-level objectives (SLOs). A single throughput number cannot answer the questions operators actually face, because generative serving couples two phases with very different service characteristics, continuous batching reshapes the effective service process, and user-visible tail latency can degrade long before aggregate throughput saturates. We present GAUSS, a workload-aware analytical and Monte-Carlo simulator that predicts the full distributions of time-to-first-token (TTFT) and time-between-tokens (TBT), together with the sustainable request rate, for a given workload, engine, and accelerator. GAUSS pairs calibrated, phase-specific latency surfaces for prefill and decode with a Lindley-recurrence model of queueing under stochastic arrivals, and estimates performance by ensemble Monte-Carlo sampling rather than by replaying a single trace. This lets GAUSS reason about workloads, routing policies, and hardware that have never been deployed, and to capture the latency tails that trace-driven simulators systematically under-sample. We describe the model and show how it supports operating-point selection, traffic segmentation, and shortest-remaining-processing-time scheduling—capacity questions that trace-driven simulators cannot address.*

## 1. Introduction

Serving large language models is now a first-order operational concern. Fleets must decide how much offered load an engine can absorb, which deployments are underutilized, and whether observed latency is dominated by prefill work, decode work, or time spent waiting in a queue. These questions cannot be answered by a single tokens-per-second figure. Generative inference proceeds in two phases—a compute-bound prefill that consumes the prompt and a memory-bound decode that emits tokens one at a time—and continuous batching dynamically mixes requests at different generation positions, so the effective service process depends on the workload itself. As a result, a deployment can look healthy on average throughput while violating first-token or inter-token latency targets for specific traffic mixes.

The difficulty is compounded by heterogeneity and burstiness. Production traffic mixes prompt lengths, cached-prefix lengths, and generated lengths, and arrives in bursts rather than at a constant rate. Capacity planning, SLO setting, utilization analysis, and cost attribution all depend on how these workload features interact with engine-specific service times and queueing dynamics under a given arrival process. What operators need is a model that is light enough to evaluate many candidate workloads, engines, and accelerators, yet rich enough to expose TTFT and TBT distributions and the request rate at which they degrade.

**Why an analytical model.** The natural alternative—replaying a recorded request trace through a discrete-event simulator—answers only a narrow question: how did one realization of one workload behave on one configuration. An analytical model instead expresses service time as a calibrated function of a request's shape (its input, cached, and output token counts) and of the engine and accelerator. Once service time is a function rather than a logged number, three capabilities follow that trace replay cannot provide:

- **Ensemble rather than time averages.** A single trace yields one time-averaged trajectory. Under the high service-time variability of LLM workloads, that trajectory is one draw from a broad distribution and systematically under-samples the tail. An analytical model can be driven by many independent stochastic realizations—an ensemble—so it estimates the entire TTFT/TBT distribution, including the P95 and P99 that SLOs are written against, rather than a single percentile estimate from one run.
- **Prediction for unseen workloads and policies.** Because service time is computed from request shape and engine calibration, the model can predict latency and capacity for workloads, routing or scheduling policies, and hardware that have never been run, and therefore appear in no trace. Trace-driven simulators are bound to the workload and configuration recorded in the trace.
- **Compactness.** Evaluating closed-form latency surfaces and a queueing recurrence is far cheaper than detailed discrete-event simulation, so an operator can sweep hundreds of candidate request rates, workloads, and engines in the time a high-fidelity simulator evaluates one.

We study the resulting prediction problem directly. Given a workload distribution expressed as weighted request buckets over input, cached, and output tokens, an arrival process with a specified mean rate, and a calibrated engine description, predict the distribution of batch sizes and backlog states, the TTFT and TBT percentile distributions, and the request rate at which tail latency degrades past a chosen threshold. This is distinct from designing a new scheduler, memory manager, or runtime: it takes the serving mechanisms established by prior work as given and asks what TTFT, TBT, and capacity will look like for a described workload, engine, and deployment.

This report makes the following contributions.

- **A workload-aware capacity model.** GAUSS represents traffic as weighted request buckets over input, cached, and output tokens, and combines this mix with a stochastic arrival process to estimate serving behavior under candidate offered loads.
- **Calibrated, phase-specific latency surfaces.** GAUSS models prefill and decode service times through engine-calibrated coefficients, exposing how context length, cached context, and batch size shape service time.
- **Monte-Carlo estimation of TTFT and TBT distributions.** GAUSS predicts latency percentiles, not only average throughput, by composing prefill work, decode-batch cost, and a Lindley-style queue-delay distribution.
- **A capacity and headroom workflow.** GAUSS sweeps candidate request rates and selects an operating point before tail latency degrades sharply, supporting capacity planning, SLO exploration, and utilization analysis.

The unifying contribution is the ability to connect workload composition, arrival burstiness, and engine calibration to the user-visible metrics that operators manage—TTFT, TBT, and sustainable request rate—for configurations that need not yet exist.

# 2. Background

GAUSS sits at the intersection of GPU execution, LLM serving runtimes, and classical queueing theory. We review each in turn: how accelerators execute the model, how batched-inference algorithms reshape the service process, and how GAUSS relates to existing simulators and capacity-prediction frameworks.

## 2.1 GPU Architecture and the Two Phases of Inference

Modern accelerators are massively parallel throughput machines: thousands of arithmetic lanes execute the same instruction across many data elements, fed by a deep memory hierarchy whose top level is a bandwidth-limited high-bandwidth memory (HBM). Two consequences govern inference performance. First, the lanes are efficient only when supplied with large, regular blocks of work; small or irregular operations leave most of the compute idle. Second, every step must read the model weights, and during generation the per-token key/value (KV) state, from HBM.

These properties split LLM inference into two regimes. Prefill processes all prompt tokens in parallel and is compute-bound: its cost grows with the number of tokens it must attend to. Decode emits one token per step and is memory-bound: each step re-reads the weights and the full KV cache to produce a single token, so the lanes are under-utilized and latency is dominated by memory traffic rather than arithmetic. The KV cache itself occupies HBM in proportion to the number of concurrently served requests and their context lengths, making memory capacity—not raw compute—the binding constraint on how many requests can share an accelerator. GAUSS abstracts exactly these effects into its calibrated latency surfaces: a context-dependent prefill cost and a batch- and context-dependent decode cost.

## 2.2 Batched Inference Algorithms for LLMs

Because a single decode step re-reads the entire model for one token, serving one request at a time wastes the accelerator. Batching amortizes each weight read across many requests, so decode throughput rises with batch size until the lanes saturate and the workload becomes compute-bound. The serving literature is, in large part, a sequence of mechanisms for keeping that batch full and well-shaped.

**Continuous batching.** Early systems formed a fixed batch and waited for every request in it to finish before admitting new work; because generated lengths vary widely, short requests finished early and their slots sat idle behind the longest requestS. Orca introduced iteration-level (continuous) batching, in which a finished request leaves and a waiting request joins after each token step, so the batch stays full without waiting for the slowest request [1].

**Paged KV-cache management.** With the batch kept busy, memory becomes the ceiling. vLLM made KV-cache memory a first-class concern through paged attention, allocating the cache in fixed-size blocks—analogous to operating-system virtual-memory pages—so that fragmentation and over-reservation no longer cap occupancy. This raises the achievable batch size, and hence throughput, on the same hardware [2].

**Chunked prefill.** A long prefill, run on its own, stalls the ongoing decode steps and produces a spike in inter-token latency. Sarathi-Serve and DeepSpeed-FastGen split a prefill into smaller chunks and interleave each chunk with a decode batch, consuming the compute slack that the memory-bound decode leaves idle. Prefill then makes progress without stalling decode, yielding a more uniform inter-token latency at a controllable throughput cost [3, 4].

GAUSS does not propose another such mechanism. It abstracts their performance consequences—the dependence of service time on context length, batch size, and chunked-prefill interleaving—into calibrated coefficients, so that an operator can predict latency and capacity for a configuration that uses these mechanisms before committing to a deployment.

## 2.3 Positioning Among Inference Simulators and Capacity Models

Several lines of work predict or simulate inference performance, and GAUSS is best understood by contrast with them.

**High-fidelity discrete-event simulators.** Vidur models schedulers, batching policies, and engine behavior in detail and can evaluate many configuration choices through discrete-event simulation [5]. Such simulators are the right tool for studying a specific scheduler or runtime at high fidelity, but they are comparatively heavy to run and are typically driven by a recorded or synthetic trace—one realization of the workload. They therefore report time-averaged behavior for that realization and are expensive to sweep across many workloads and rates.

**Profile-driven capacity frameworks.** InferLine and Clockwork established that measured per-stage profiles, latency SLOs, and arrival assumptions can drive SLO-aware provisioning, and that predictable execution is a prerequisite for tight capacity reasoning [8, 9]. These frameworks predate the autoregressive, two-phase structure that dominates LLM serving and treat inference as a pipeline of fixed-cost stages rather than a context- and batch-dependent service process.

**Workload characterization and queueing foundations.** BurstGPT documents that real LLM traffic is heterogeneous in prompt and response lengths and bursty in arrivals, motivating workload-aware prediction over single-number throughput summaries [7]. On the theory side, Lindley’s recurrence supplies the standard description of waiting time in a single-server queue with general service and arrival distributions [11]; Cox’s analysis of point processes underpins stochastic arrival modeling [12]; and modern treatments of M/G/1 queues, heavy-tailed job sizes, and size-based scheduling provide the analytical tools we draw on [13]. Recent work has also analyzed scheduling policies for LLM serving under heterogeneous service times [10].

GAUSS exists at the intersection of these theoretical models, benchmark/profile driven frameworks and inference simulators: lighter and more flexible than a discrete-event simulator and richer than a static benchmark. It combines calibrated, phase-specific analytical latency surfaces with Monte-Carlo sampling over workload buckets and stochastic arrivals to produce TTFT, TBT, and sustainable-rate predictions. Crucially, because it is analytical and probabilistic rather than trace-driven, it estimates ensemble distributions—capturing tails—and extrapolates to workloads, policies, and hardware absent from any trace. No prior work we are aware of combines calibrated phase-specific latency surfaces, weighted workload buckets, stochastic arrivals, and a queue-backlog recurrence into a single compact predictor of TTFT, TBT, and capacity.

# 3. Design of the Simulator

This section gives a mathematical overview of how GAUSS turns a workload description and an engine calibration into latency distributions. We describe the model for decoder-only models and a single serving stage. We first model the service time of an individual request, then the queueing delay induced by stochastic arrivals, and finally compose the two into TTFT and TBT distributions. Throughout, lowercase Latin and Greek letters denote calibrated or sampled scalars; they are introduced where used.

## 3.1 Workload and Arrival Model

A workload is represented as a discrete distribution over request "buckets." Bucket j is a triple of token counts—input length $s_j$, cached (already-resident prefix) length $c_j$, and output length $o_j$—carried with probability $w_j$, where the weights sum to one. Empirically, during busy periods this distribution is stable across days because a large shared system prompt is cached and reused by most requests, which concentrates context lengths near the system-prompt length and lets the mix be fit once and reused for planning. Requests arrive as a Poisson process of mean rate λ i.e. the inter-arrival time A between successive requests is exponential with mean 1/λ.

## 3.2 Prefill Service Time

Under chunked prefill, a request's prompt is processed in chunks of a fixed size Δ. A request whose input length is s and whose cached prefix length is c must process its uncached tokens, so the number of chunks is

$$n = \lceil (s - c) / \Delta \rceil .$$

The i-th chunk attends to all tokens resident when it runs—the cached prefix plus the i chunks already processed—so its attended context is $g_i = c + i\Delta$. Calibration shows the per-chunk latency to be affine in this attended context, with engine coefficients a and b:

$$\ell_i = a + b \cdot g_i = a + b(c + i\Delta) .$$

Summing over the n chunks gives the prefill compute of the request. Because the attended context grows linearly with the chunk index, the sum is an arithmetic series, and the total is quadratic in the number of chunks:

$$L_pre(s, c) = \Sigma_i \ell_i = a \cdot n + b \cdot \sigma , \quad \sigma = \Sigma_i g_i = n \cdot c + \Delta \cdot n(n+1)/2 .$$

This closed form is the analytical counterpart of the empirical observation that per-chunk latency is linear in the resident cache while the per-request total grows quadratically with the number of chunks. It lets GAUSS compute a request's prefill cost directly from its shape, with no need to execute it.

## 3.3 Decode-Batch Latency

During generation, a batch of B requests with aggregate resident context C advances by one token per step. Because decode is memory-bound, the per-step latency is well described by an affine surface in the batch size and the average context per request, with calibrated coefficients p, q, and r:

$$d(B, C) = p + q \cdot (C / B) + r \cdot B .$$

The C/B term captures the cost of reading each request's KV state, and the linear B term captures the marginal cost of widening the batch. In practice the coefficients are piecewise across batch-size regimes; we write a single affine form for clarity. The decode-batch latency couples back into prefill service time, because chunked prefill folds each prefill chunk into a concurrent decode batch.

### 3.4 Service Time Under Concurrent Prefill and Decode

When prefills and decodes are served concurrently, a prefill chunk does not run in isolation: it rides along with the decode batch active at that step. A request's prefill service time therefore comprises its own chunked-prefill compute together with the decode-batch cost of the steps its chunks occupy. Writing $\bar{d}$ for the decode-batch latency of the concurrent batch and $\eta$ for an engine efficiency factor that converts modeled work into wall-clock time, the prefill service time of a request is

$$S(s, c) = ( a \cdot n + b \cdot \sigma + n \cdot \bar{d} ) / \eta .$$

The same interleaving determines inter-token latency during the steady decode phase: each decode step costs $d(B, C)$, and folding a prefill chunk into a step inflates that step by the chunk's marginal cost. Averaging over the generated tokens yields the per-request TBT. Holding the chunk size $\Delta$ small keeps this inflation bounded, which is the analytical statement of the chunked-prefill latency–throughput trade-off.

### 3.5 Service Rate and the Operating Point

The expected service time of a request drawn from the workload, and the corresponding service rate $\mu$, follow by averaging over the buckets:

$$E[S] = \Sigma_j \, w_j \cdot S(s_j, c_j) , \quad \mu = 1 / E[S] .$$

In steady state throughput equals the arrival rate, so the system is stable only while the load $\rho = \lambda/\mu$ is below one; the maximum sustainable request rate is $\lambda^* = \mu$. Because GAUSS computes $\mu$ analytically from the latency coefficients and the workload mix, it can report the throughput ceiling for any workload—including hypothetical ones—without measuring it.

### 3.6 Queueing Delay via a Lindley Recurrence

We model admission to the (chunked-)prefill server as a single-server first-in, first-out queue with general service times—an M/G/1 abstraction. Let $S_k$ be the service time of the k-th request, sampled from the workload, and $A_k$ its inter-arrival time, sampled as exponential with rate $\lambda$. The waiting time obeys Lindley's recurrence,

$$W_{k+1} = \max( 0, W_k + S_k - A_k ) .$$

Rather than solve for a stationary distribution in closed form, GAUSS evaluates the recurrence by Monte-Carlo over many sampled service and inter-arrival sequences, yielding the full ensemble distribution of the queue delay W rather than only its mean. This is the step that turns calibrated service times and a stochastic arrival rate into a latency distribution.

### 3.7 Composing TTFT and TBT

The first token of a request appears only after it has waited in the queue and then finished its prefill. TTFT is therefore the sum of a queue-delay sample and a prefill-service sample:

$$TTFT = W + S(s, c)$$

Sampling W from the Lindley ensemble and S from the workload, and adding them, gives the steady-state TTFT distribution. TBT is obtained from the decode-batch latency together with the interleaved-prefill inflation of Section 3.4, averaged over the request's generated tokens. Sweeping the arrival rate λ and the workload mix traces out how the TTFT and TBT percentiles move, and locates the request rate at which the tail degrades sharply.

## 3.8 Batch Size, Backlog, and the Estimation Procedure

The batch size is not a free parameter: in steady state it is determined by how much work fits in the per-request time budget set by the offered load. Balancing prefill and decode work against that budget yields a quadratic equation whose root is the operating batch size, which GAUSS solves in closed form, and so the simulator reports distributions over batch sizes alongside the latency percentiles. Algorithm 1 summarizes the procedure.

**Algorithm 1. GAUSS Monte-Carlo estimation (single stage, decoder-only)**

```
Input:  workload buckets {(s_j, c_j, o_j, w_j)}; arrival rate λ;
        engine coefficients (a, b, p, q, r, η, Δ); sample counts M, N
Output: distributions of TTFT, TBT; service rate μ; sustainable rate λ*

1  for each bucket j:                                  // service times
      n_j   = ceil((s_j - c_j) / Δ)
      σ_j   = n_j*c_j + Δ*n_j*(n_j+1)/2
      S_j   = (a*n_j + b*σ_j + n_j*d_bar) / η
2  E[S] = Σ_j w_j * S_j ;   μ = 1 / E[S] ;   λ* = μ
3  W = 0 ;  Q = [ ]                                    // Lindley sweep
      for k = 1 .. M:
         draw bucket ~ w ;  S_k = S_bucket ;  A_k ~ Exp(λ)
         W = max(0, W + S_k - A_k) ;  append W to Q
4  for i = 1 .. N:                                     // TTFT samples
         draw bucket ~ w ;  S = S_bucket ;  W ~ Q
         TTFT_i = W + S
5  for i = 1 .. N:                                     // TBT samples
         form decode batch (size B, context C) for this load
         TBT_i = d(B, C)/η inflated by interleaved prefill chunks
6  return percentiles of {TTFT_i}, {TBT_i}, and μ, λ*
```

### *Modeling assumptions*

The model rests on a small number of assumptions, stated here so they can be revisited as workloads and engines evolve.

- **Stationarity at peak.** The workload distribution is treated as stable over the planning window, justified by the cached-system-prompt concentration described in Section 3.1.
- **Poisson arrivals.** Requests arrive as a Poisson process of mean rate λ, so inter-arrival times are exponential.
- **Single-server prefill admission.** Queueing is modeled as one M/G/1 server whose "job" is a request's full chunked prefill served first-in, first-out; the Lindley recurrence follows directly.
- **Calibrated latency surfaces.** Per-chunk prefill latency is affine in attended context, and decode-batch latency is (piecewise) affine in batch size and average context; coefficients are fit per engine and accelerator.

## 4. Example Use Cases

Since GAUSS predicts service-time and latency distributions analytically, it answers operational questions that a trace-driven simulator cannot. We illustrate three concrete tasks from the tutorial and then a forward-looking class of applications enabled by ensemble estimation.

### 4.1 Operating-Point and Capacity Calculation

Throughput in steady state equals the offered arrival rate, and the maximum sustainable rate is the service rate $\lambda^* = \mu = 1/E[S]$. GAUSS computes $\mu$ directly from the latency coefficients and the workload mix, then sweeps the arrival rate to trace the TTFT percentile curve. That curve has a sharp knee: as $\lambda$ approaches $\mu$ the queue delay grows without bound and the tail explodes. Operating just below the knee captures most of the achievable utilization while keeping the tail within SLO, and the gap between a chosen operating point and the knee is exactly the headroom available for routing and provisioning.

**Why prior approaches fall short.** A trace-driven simulator would need a representative trace at every candidate rate and for every workload mix and could only report the realized metrics for those fixed traces; it cannot return $\mu$ analytically, and it cannot evaluate a workload mix it has not recorded. GAUSS returns both the ceiling and the full TTFT curve for any mix, including ones the operator is only contemplating.

### 4.2 Traffic Segmentation

Prefill service times are heavy-tailed, because context lengths vary widely. In an M/G/1 queue the mean waiting time grows with the second moment of service time,

$$E[W] = \lambda \cdot E[S^2] \,/\, (\, 2(1 - \rho)\,)\ ,$$

so a few very long requests inflate the delay of everyone behind them—the head-of-line blocking that the inspection paradox makes precise: a random arrival is disproportionately likely to land behind a long request already in service. Splitting traffic into segments by context length and serving each segment from its own pool reduces the within-segment service variability, and hence the waiting time. The leverage point is choosing the segment boundary, which requires knowing $E[S]$ and $E[S^2]$ on either side of any candidate split.

**Why prior approaches fall short.** GAUSS computes the service-time moments of each candidate segment analytically from the workload buckets and the latency surfaces, so it can search for the boundary that minimizes overall delay. A trace-driven simulator would have to re-run the split workload for each candidate boundary, and could not evaluate a partition of a workload it has not observed.

### 4.3 Shortest-Remaining-Processing-Time Scheduling

Serving the request with the shortest remaining work first (SRPT) minimizes mean response time and is the canonical antidote to head-of-line blocking under heavy-tailed service [13]. SRPT requires, at every decision point, the remaining processing time of each waiting request. For chunked prefill this is exactly the cost of the chunks a request has yet to process, which the analytical model gives in closed form: the remaining time of a request with $c'$ tokens still resident and a tokens-worth of context per chunk is the same affine-in-context sum used in Section 3.2, evaluated over its remaining chunks.

**Why prior approaches fall short.** Without an analytical service-time model, a scheduler must estimate remaining processing time from coarse heuristics—token counts or running averages—which are exactly the signals that mislead under heavy tails. GAUSS supplies the remaining-time function that SRPT needs, and lets one quantify, before deployment, how much tail-latency reduction the policy would deliver for a given workload.

### 4.4 Future Use Cases: Ensemble Estimation for Fleet Operations

The capability that underlies all of the above—and that distinguishes GAUSS most sharply from trace replay—is ensemble estimation. A trace-driven simulator produces a time average: one long trajectory whose tail is a single, noisy draw. GAUSS produces an ensemble: many independent stochastic realizations of the same operating condition, from which the full distribution, and therefore the P95 and P99, can be read directly. Under the high variability and burstiness of real LLM traffic, the two differ materially, and the tail percentiles that SLOs target are precisely where a single trajectory is least reliable. Only an analytical, probabilistic model can generate the ensemble cheaply, because it samples service times from a calibrated function rather than replaying recorded events.

This unlocks a broader class of fleet-level applications. Reliable tail estimates per deployment make it possible to route and load-balance requests toward the instance with the most headroom under the current mix, to provision capacity against a target percentile rather than an average, and to run what-if analyses across engines and accelerators before any of them is deployed. In each case the operative requirement is a trustworthy latency distribution for a configuration that may not yet exist—the regime in which GAUSS, rather than a trace-driven simulator, is the realistic tool.

## References

[1] G.-I. Yu et al. *Orca: A Distributed Serving System for Transformer-Based Generative Models*. Proc. USENIX OSDI, 2022.

[2] W. Kwon et al. *Efficient Memory Management for Large Language Model Serving with PagedAttention*. Proc. ACM SOSP, 2023. arXiv:2309.06180.

[3] A. Agrawal et al. *Taming Throughput-Latency Tradeoff in LLM Inference with Sarathi-Serve*. arXiv:2403.02310, 2024.

[4] C. Holmes et al. *DeepSpeed-FastGen: High-throughput Text Generation for LLMs via MII and DeepSpeed-Inference*. arXiv:2401.08671, 2024.

[5] B. Wu et al. *Fast Distributed Inference Serving for Large Language Models*. arXiv:2305.05920, 2023.

[6] A. Agrawal et al. *Vidur: A Large-Scale Simulation Framework for LLM Inference*. arXiv:2405.05465, 2024.

[7] Y. Wang et al. *BurstGPT: A Real-world Workload Dataset to Optimize LLM Serving Systems*. arXiv:2401.17644, 2024.

[8] D. Crankshaw et al. *InferLine: ML Inference Pipeline Composition Framework*. Proc. USENIX NSDI, 2020.

[9] A. Gujarati et al. *Clockwork: Predictable Low Latency for Deep Learning Inference*. Proc. USENIX OSDI, 2020.

[10] A. Bari, P. Hegde, and G. de Veciana. *Optimal Scheduling Algorithms for LLM Inference: Theory and Practice*. Proc. ACM on Measurement and Analysis of Computing Systems (POMACS), 2025.

[11] D. V. Lindley. *The Theory of Queues with a Single Server*. Mathematical Proceedings of the Cambridge Philosophical Society, 1952.

[12] D. R. Cox. *Some Statistical Methods Connected with Series of Events*. Journal of the Royal Statistical Society: Series B, 1955.

[13] M. Harchol-Balter. *Performance Modeling and Design of Computer Systems: Queueing Theory in Action*. Cambridge University Press, 2013.